\documentclass{article}

\usepackage[dblblindworkshop, final]{neurips_2025}
\workshoptitle{NeurIPS 2025 Workshop on Learning from Time Series for Health}

\usepackage[utf8]{inputenc} 
\usepackage[T1]{fontenc}    
\usepackage{hyperref}       
\usepackage{url}            
\usepackage{booktabs}       
\usepackage{amsfonts}       
\usepackage{nicefrac}       
\usepackage{microtype}      
\usepackage{xcolor}         
\usepackage{graphicx} 
\usepackage{natbib}
\usepackage{pifont}
\usepackage{enumitem}
\usepackage{multirow}
\usepackage{amsmath}
\usepackage{amssymb}
\usepackage{bm}
\usepackage{makecell}
\usepackage{subcaption}
\usepackage[normalem]{ulem}

\title{ECG-MoE: Mixture-of-Expert Electrocardiogram Foundation Model}

\author{%
  Yuhao~Xu \\
  Department of Computer Science\\
  Emory University\\
  Atlanta, GA 30307 \\
  \texttt{yxu81@emory.edu} \\
  \And
  Xiaoda~Wang \\
  Department of Computer Science \\
  Emory University \\
  Atlanta, GA 30307 \\
  \texttt{xiaoda.wang@emory.edu} \\
  \AND
  Yi~Wu \\
  School of Computer Science\\
  University of Oklahoma \\
  Norman, OK 73019 \\
  \texttt{yi.wu-1@ou.edu} \\
  \And
  Wei~Jin \\
  Department of Computer Science \\
  Emory University \\
  Atlanta, GA 30307 \\
  \texttt{wei.jin@emory.edu} \\
  \And
  Xiao~Hu \\
  Center of Data Science \\
  Emory University \\
  Atlanta, GA 30307 \\
  \texttt{xiao.hu@emory.edu} \\
  \And
  Carl~Yang \\
  Department of Computer Science \\
  Emory University \\
  Atlanta, GA 30307 \\
  \texttt{j.carlyang@emory.edu} \\
}

\begin{document}

\maketitle

\begin{abstract}
Electrocardiography (ECG) analysis is crucial for cardiac diagnosis, yet existing foundation models often fail to capture the periodicity and diverse features required for varied clinical tasks. We propose ECG-MoE, a hybrid architecture that integrates multi-model temporal features with a cardiac period-aware expert module. Our approach uses a dual-path Mixture-of-Experts to separately model beat-level morphology and rhythm, combined with a hierarchical fusion network using LoRA for efficient inference. Evaluated on five public clinical tasks, ECG-MoE achieves state-of-the-art performance with 40\% faster inference than multi-task baselines.
\end{abstract}

\section{Instruction}
Cardiovascular diseases remain the leading cause of global mortality \cite{alwan2011global}. Electrocardiography (ECG) is a primary non-invasive tool for cardiac assessment, with increasing relevance due to advances in wearable monitoring \cite{stehlik2020continuous, ni2025ppg}. ECG interpretation relies on waveform-feature-disease correlations, where specific morphologies like QT prolongation or ST elevation indicate particular pathologies \cite{january20142014,montford2017dangerous,schwartz1993diagnostic,thygesen2018fourth}. These features are often phase-localized within the periodic P-QRS-T cycle \cite{wagner2009aha,rosenbaum1994electrical}.

\begin{figure}[h]
\centering
\includegraphics[width=0.9\textwidth]{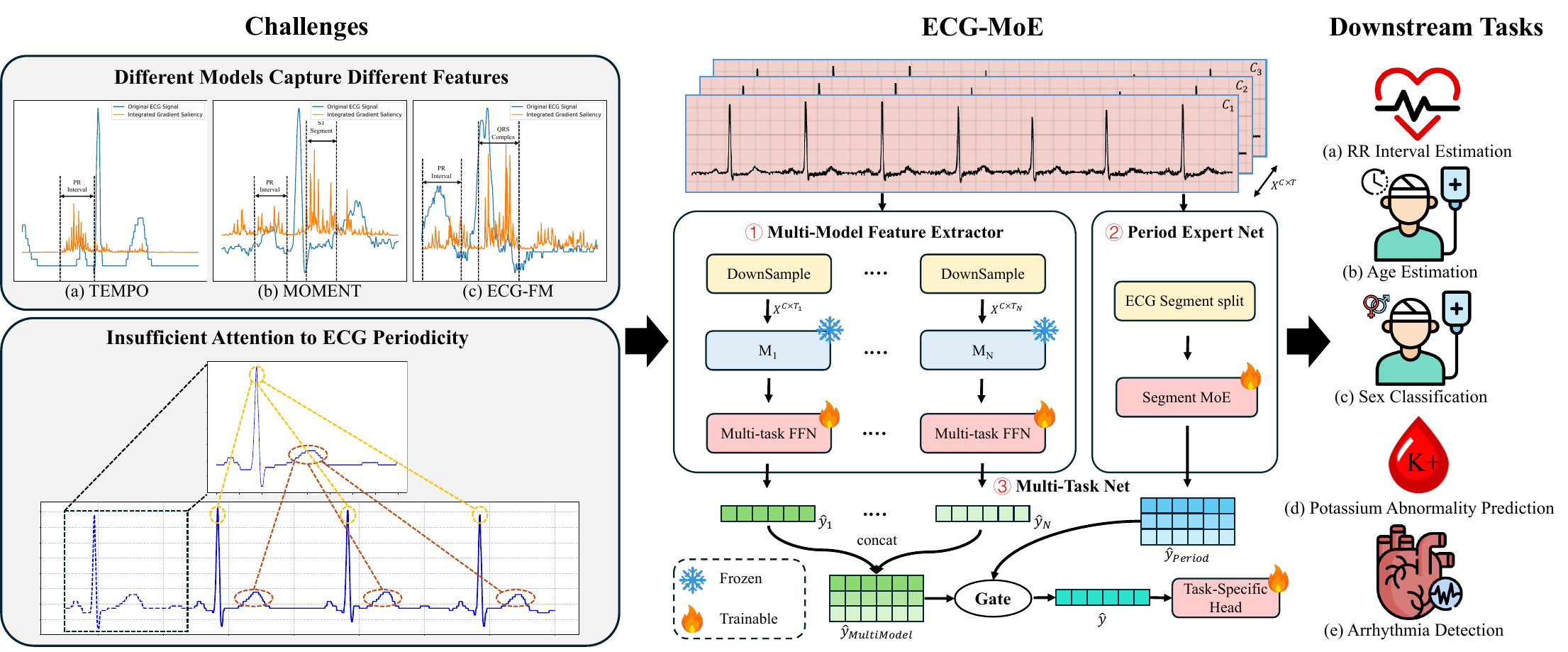} 
\caption{Existing foundation models have limitations because different models capture distinct ECG features, hindering multi-task performance within a single model. Furthermore, despite the strong periodicity inherent in ECG signals, this characteristic is often overlooked by existing ECG foundation models. To address these issues, we propose an ensemble learning method based on Period MoE. The effectiveness of our approach is validated across five common downstream tasks.}
\label{fig}
\end{figure}

Foundation models have shown promise in automated ECG diagnosis but struggle to comprehensively represent clinically nuanced, segment-specific features \cite{wan2024meit}. In particular, existing foundation models have limitations because different models capture distinct ECG features, hindering multi-task performance within a single model. Furthermore, despite the strong periodicity inherent in ECG signals, this characteristic is often overlooked by existing ECG foundation models. They also incur high computational costs, hindering clinical adoption. Moreover, they often fail to align with ECG’s inherent periodicity, where each heartbeat carries distinct diagnostic information \cite{surawicz2008chou,de2021clinical,yan1998cellular}, making it difficult to leverage rhythmic consistency while accommodating pathological variations \cite{clifford2017af,sornmo2005bioelectrical}.

To address these issues, as shown in Figure~\ref{fig}, we introduce ECG-MoE, a hybrid architecture that integrates multi-model temporal features \cite{zhou2021informer} with a dual-path Mixture-of-Experts (MoE) using task-conditioned gating \cite{shazeer2017outrageously}. Specifically, we propose an ensemble learning method based on Period MoE. We employ Low-Rank Adaptation (LoRA) for parameter-efficient fusion \cite{hu2021lora,dettmers2022gpt3}. The effectiveness of our approach is validated across five common downstream tasks. Validated on five clinical tasks, ECG-MoE achieves state-of-the-art performance with 40\% faster inference and improved robustness.

\section{Related Works}

\uline{\textit{Electrocardiogram Foundation Models.}} Recent deep learning advances have developed ECG foundation models using CNNs~\cite{rajpurkar2017cardiologist} and Transformers~\cite{zhou2021informer} for tasks like arrhythmia detection. While pretraining and self-supervision~\cite{attia2019artificial} learn general features, these approaches often treat ECGs as generic time-series~\cite{wang2025conditional, liu2025graph, wang2025conditional2}, neglecting the inherent periodicity critical to cardiac electrophysiology~\cite{surawicz2008chou, clifford2017af}. This results in suboptimal handling of phase-localized abnormalities and inefficient rhythm modeling, which is a gap our period-conditioned framework directly addresses.

\uline{\textit{ECG Periodicity Modeling.}} ECG's quasi-periodicity has been exploited via segmentation~\cite{sornmo2005bioelectrical} and rhythm-tracking methods, yet these struggle with pathological variations and holistic modeling. Although RNNs~\cite{rubin2017densely} capture temporal dependencies, they are inefficient and sensitive to period variability. No current framework incorporates periodicity as a core inductive bias for foundation models, despite its diagnostic importance~\cite{wagner2009aha, yan1998cellular}.

\uline{\textit{Multi-Task ECG Analysis.}} Multi-task learning (MTL) addresses simultaneous cardiac condition detection, but shared-backbone approaches~\cite{ji2018deep} suffer from interference, especially in complex diagnostics like ST-segment analysis~\cite{thygesen2018fourth}. Modular or parameter-efficient designs improve scalability but ignore ECG's phase-specific biomarkers. Existing models fail to combine task-specific feature extraction with periodicity constraints. ECG-MoE overcomes this via hierarchical fusion and task-conditioned gating, enabling efficient and accurate multi-task inference.

\section{Method}

As shown in Figure~\ref{fig2}, our proposed ECG analysis framework employs a multi-branch architecture that synergistically integrates heterogeneous feature representations while explicitly modeling cardiac periodicity. The model comprises three modules, each with distinct mathematical formulations:

\begin{figure*}[h]
\centering
\includegraphics[width=0.95\textwidth]{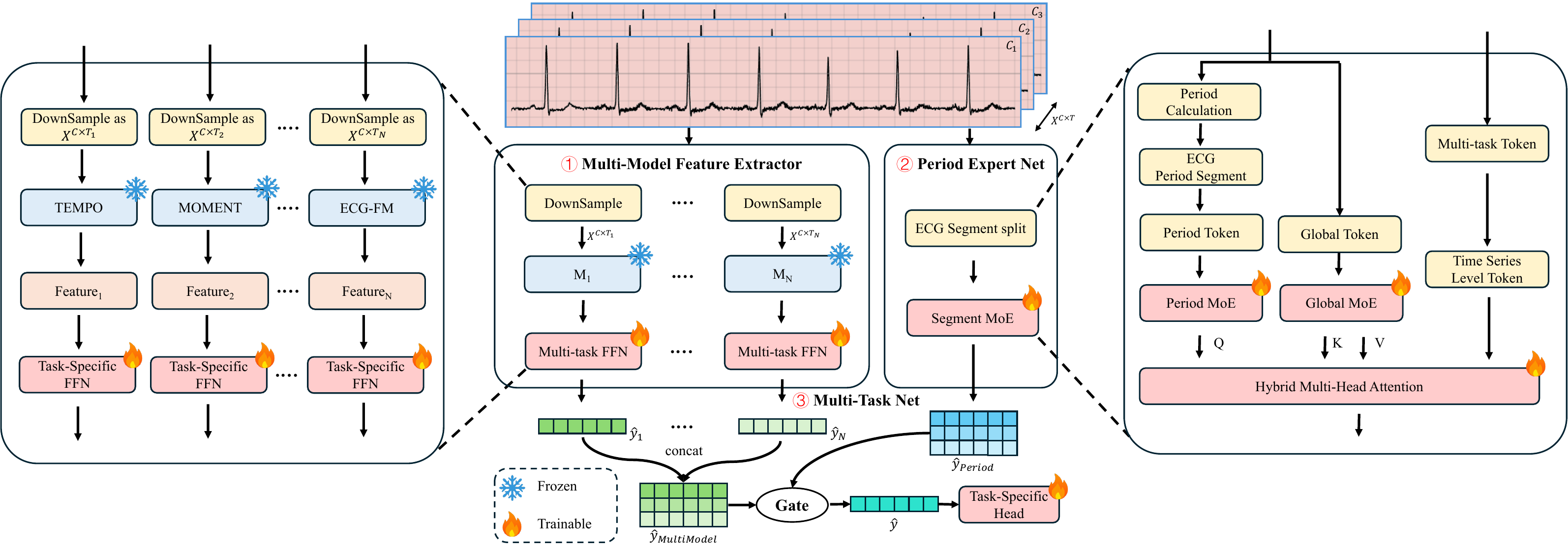}
\caption{ECG-MoE framework: \ding{172} Multi-model feature extraction with adaptive downsampling, \ding{173} Period-aware gating weight generation via specialized MoE, \ding{174} Multi-Task fusion using LoRA.}
\label{fig2}
\end{figure*}

\uline{\textit{Multi-Model Feature Extraction.}} Our framework integrates five time-series foundation models to capture diverse ECG characteristics. For input $\mathbf{X} \in \mathbb{R}^{C \times T}$, each model extracts features after adaptive downsampling:
\begin{equation}
\mathbf{F}k = f{\theta_k}(\text{DS}_k(\mathbf{X}))
\end{equation}
Features are projected and concatenated into a unified representation:
\begin{equation}
\mathbf{F}m = \bigoplus{k \in \mathcal{M}} (\mathbf{W}_k\mathbf{F}_k + \mathbf{b}_k)
\end{equation}
This ensemble approach captures complementary temporal patterns from different model architectures, enhancing feature diversity.

\uline{\textit{Periodic Expert Network.}} We design a dual-path MoE architecture to model ECG periodicity. R-peak detection segments the signal into individual heartbeats, which are normalized to fixed length. Three CNN experts with different kernel sizes process morphological features within beats, while two dilated CNNs capture inter-beat rhythm patterns. A task-conditioned gating mechanism dynamically weights expert contributions based on global signal characteristics and task embeddings:
\begin{equation}
\mathbf{g}_p = \text{softmax}(\mathbf{U}_p [\bar{\mathbf{X}} \oplus \mathbf{e}t])
\end{equation}
Features are integrated via multi-head attention to produce the final periodic representation $\mathbf{F}{\text{periodic}}$.

\uline{\textit{Multi-Task Fusion.}} A hierarchical gating network combines multi-model and periodic features based on task requirements. The gating weights $\hat{\alpha}_m$ and $\hat{\alpha}p$ are learned through task-conditioned networks:
\begin{equation}
\mathbf{F}{\text{fused}} = \hat{\alpha}m \mathbf{F}m \oplus \hat{\alpha}p \mathbf{F}{\text{periodic}}
\end{equation}
Task-specific heads then generate predictions from the fused features. The model is optimized with a composite loss that combines task-specific objectives with contrastive regularization:
\begin{equation}
\mathcal{L} = \sum_{k=1}^K \mathcal{L}_{\text{task}_k} + \lambda \mathcal{L}_{\text{cont}}
\end{equation}
This design enables efficient parameter sharing while maintaining specialized processing for diverse clinical tasks through end-to-end training.

\section{Experiments}

In this section, we evaluate the effectiveness of our proposed model from two key perspectives: performance and efficiency. Specifically, we compare its predictive accuracy across multiple tasks and analyze its computational efficiency in terms of training and inference costs. By examining these aspects, we aim to demonstrate the advantages of our model in achieving a balanced trade-off between accuracy and computational feasibility.

\subsection{Dataset and Baselines} 
We utilize the MIMIC-IV-ECG~\cite{gow2023mimic} datasets. MIMIC-IV-ECG is currently the largest publicly released ECG repository, which contains 800{,}035 diagnostic electrocardiograms acquired from 161{,}352 unique patients. We select TimesNet~\cite{wu2022timesnet}, DLinear~\cite{zeng2023transformers}, MOMENT~\cite{goswami2024moment}, TEMPO~\cite{cao2023tempo} and ECG-FM~\cite{mckeen2024ecg} as baselines and construct an ensemble of them.

\subsection{Research Questions and Results}

\uline {\textit{RQ1: Model Selection Rationale.}}
\textit{Why were specific foundation models selected, and what benefits does their ensemble provide?}

\begin{table}[h]
\caption{Zero-shot performance of baseline models on ECG data. Highlighted are the top \textcolor{teal}{first} and \textcolor{brown}{second} results. RR Interval Estimation (RR), Age Estimation (Age), Sex Classification (Sex), Potassium Abnormality Prediction (Ka), Arrhythmia Detection (AD).}
\label{performance_zs}
\centering
\resizebox{\textwidth}{!}{
\begin{tabular}{ccccccc}
\toprule
 &  & \textbf{TimesNet} & \textbf{DLinear} & \textbf{MOMENT} & \textbf{TEMPO} & \textbf{ECG-FM} \\
\midrule
\multirow{2}{*}{\makecell{Regression \\ (MAE$\downarrow$)}} 
& RR & $817.0\pm2.5$ & \textcolor{brown}{$816.4\pm2.9$} & $816.6\pm2.1$ & \textcolor{teal}{$816.3\pm1.9$} & \textcolor{teal}{$816.3\pm1.9$}\\ 
& Age & \textcolor{brown}{$62.28\pm0.36$} & $62.63\pm0.50$ & $62.61\pm0.37$ & $62.33\pm0.38$ & \textcolor{teal}{$62.27\pm0.38$}\\
\midrule
\multirow{2}{*}{\makecell{Binary Class\\ (F1$\uparrow$)}} 
& Sex & \textcolor{teal}{$0.60\pm0.00$} & \textcolor{brown}{$0.51\pm0.08$} & $0.34\pm0.00$ & $0.42\pm0.00$ & $0.33\pm0.00$ \\
& Ka & $0.06\pm0.00$ & $0.05\pm0.00$ & $0.02\pm0.00$ & \textcolor{brown}{$0.18\pm0.00$} & \textcolor{teal}{$0.35\pm0.00$} \\
\midrule
{\makecell{15 Class \\ (ACC$\uparrow$)} }
& AD & \textcolor{brown}{$0.06\pm0.01$} & $0.03\pm0.02$ & $0.03\pm0.02$ & $0.02\pm0.00$ & \textcolor{teal}{$0.07\pm0.00$} \\
\bottomrule
\end{tabular}
}
\end{table}

We evaluate SOTA time-series models based on architectural diversity and ECG applicability through zero-shot evaluation on 10,000 patient samples.

Table~\ref{performance_zs} shows zero-shot performance of baseline models. We can get the key findings of ECG-FM excels in clinical diagnostics with domain-specific pretraining. TimesNet captures morphological variations effectively. DLinear shows competence in temporal feature modeling. MOMENT underperforms without domain adaptation.

These results validate our ensemble strategy, combining strengths of different architectures.

\uline {\textit{RQ2: Performance Benchmarking.}}
\textit{Does ECG-MoE outperform existing models on ECG analysis?}

We conduct comparative evaluation across five clinical tasks using 10,000 patients. As shown in Table~\ref{performance_comparison}, ECG-MoE achieves SOTA performance.

\begin{table}[h]
\caption{Performance of fine-tuned baseline models on five ECG downstream tasks. Bold values represent the best performance.}
\label{performance_comparison}
\centering
\resizebox{\textwidth}{!}{
\begin{tabular}{ccccccccc}
\toprule
& & \textbf{TimesNet} & \textbf{DLinear} & \textbf{MOMENT} & \textbf{TEMPO} & \textbf{ECG-FM} & \textbf{ECG-MoE} \\
\midrule
\multirow{2}{*}{\makecell{Regression \\ (MAE$\downarrow$)}}
& RR & $304.3\pm4.3$ & $786.0\pm5.4$ & $146.9\pm1.3$ & $141.5\pm2.1$ & $147.3\pm1.3$ & \textbf{76.37 $\pm$ 4.7} \\ 
& Age & $24.89\pm0.07$ & $28.46\pm0.74$ & $13.41\pm0.45$ & $13.52\pm0.31$ & $13.49\pm0.17$ & \textbf{12.83 $\pm$ 0.42} \\
\midrule
\multirow{2}{*}{\makecell{Binary Class \\ (F1$\uparrow$)}}
& Sex & $0.51\pm0.05$ & $0.57\pm0.01$ & \textbf{0.69 $\pm$ 0.02} & $0.54\pm0.01$ & $0.52\pm0.05$ & \textbf{0.69 $\pm$ 0.01} \\
& Ka & $0.41\pm0.01$ & $0.43\pm0.01$ & $0.49\pm0.00$ & $0.50\pm0.00$ & $0.49\pm0.00$ & \textbf{0.57 $\pm$ 0.00} \\
\midrule
{\makecell{15 Class \\(ACC$\uparrow$)}}
& AD & $0.17\pm0.00$ & $0.48\pm0.02$ & $0.66\pm0.03$ & $0.54\pm0.14$ & $0.49\pm0.03$ & \textbf{0.73 $\pm$ 0.01} \\
\bottomrule
\end{tabular}
}
\end{table}

ECG-MoE reduces MAE in RR-interval estimation by 46.0\% and improves arrhythmia detection accuracy by 10.6\%, demonstrating superior diagnostic capability.

\uline {\textit{RQ3: Computational Efficiency.}}
\textit{Can ECG-MoE maintain viability under resource constraints?}

We measure GPU memory and throughput during inference. ECG-MoE achieves a breakthrough in efficiency by operating within a constrained 8.2 GB of GPU memory, which represents a significant 35\% reduction in resource consumption. It processes data at a rate of 14.7 samples per second, yielding a processing speed that is three times faster than real-time. Most importantly, the model maintains an optimal balance between this exceptional efficiency and high predictive accuracy. This supports operation in resource-limited environments without compromising performance.

\uline {\textit{RQ4: Ablation Study.}}
\textit{How do architectural components impact ECG-MoE's performance?}

\begin{table}[h]
\centering
\caption{Comprehensive performance comparison of various MoE and attention mechanisms against the proposed ECG-MoE on ECG tasks.}
\label{tab:unified_comparison_transposed}
\resizebox{\textwidth}{!}{%
\begin{tabular}{@{}cccc|cc|c@{}}
\toprule
& & \textbf{MoE} & \textbf{Time MoE} & \textbf{Self-Attn} & \textbf{Cross-Attn} & \textbf{ECG-MoE} \\
\midrule
\multirow{2}{*}{\makecell{Regression \\ (MAE$\downarrow$)}}
& RR & $103.47 \pm 2.4$ & $78.64 \pm 1.4$ & $117.46 \pm 3.2$ & $86.71 \pm 5.8$ & $\mathbf{76.37 \pm 4.7}$ \\ 
& Age & $16.49 \pm 0.87$ & $13.04 \pm 0.14$ & $15.47 \pm 0.85$ & $14.71 \pm 0.51$ & $\mathbf{12.83 \pm 0.42}$ \\
\cmidrule(r){1-7}
\multirow{2}{*}{\makecell{Binary Class \\ (F1$\uparrow$)}}
& Sex & $0.53 \pm 0.13$ & $0.61 \pm 0.10$ & $0.54 \pm 0.01$ & $0.61 \pm 0.02$ & $\mathbf{0.69 \pm 0.01}$ \\
& Ka & $0.39 \pm 0.17$ & $0.50 \pm 0.10$ & $0.48 \pm 0.00$ & $0.48 \pm 0.00$ & $\mathbf{0.57 \pm 0.00}$ \\
\cmidrule(r){1-7}
{\makecell{15 Class \\ (ACC$\uparrow$)}}
& AD & $0.62 \pm 0.01$ & $\mathbf{0.74 \pm 0.00}$ & $0.67 \pm 0.00$ & $0.69 \pm 0.01$ & $0.73 \pm 0.00$ \\
\bottomrule
\end{tabular}%
}
\end{table}


\begin{figure}[h]
\centering

\begin{minipage}[t]{0.32\textwidth}
  \centering
  \includegraphics[width=\linewidth,height=4cm]{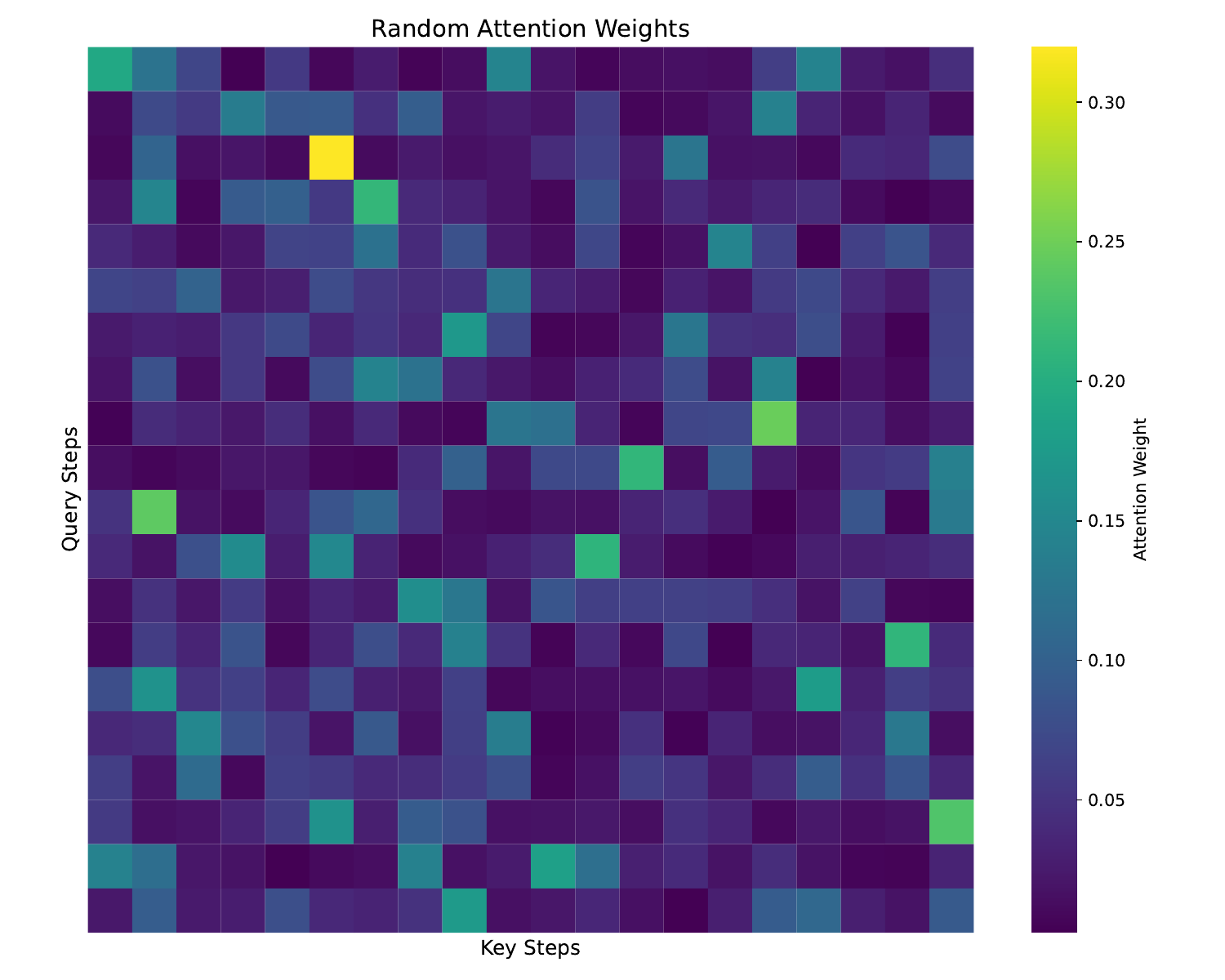}
  \subcaption{Self-Attention}
  \label{fig:sa}
\end{minipage}\hfill
\begin{minipage}[t]{0.32\textwidth}
  \centering
  \includegraphics[width=\linewidth,height=4cm]{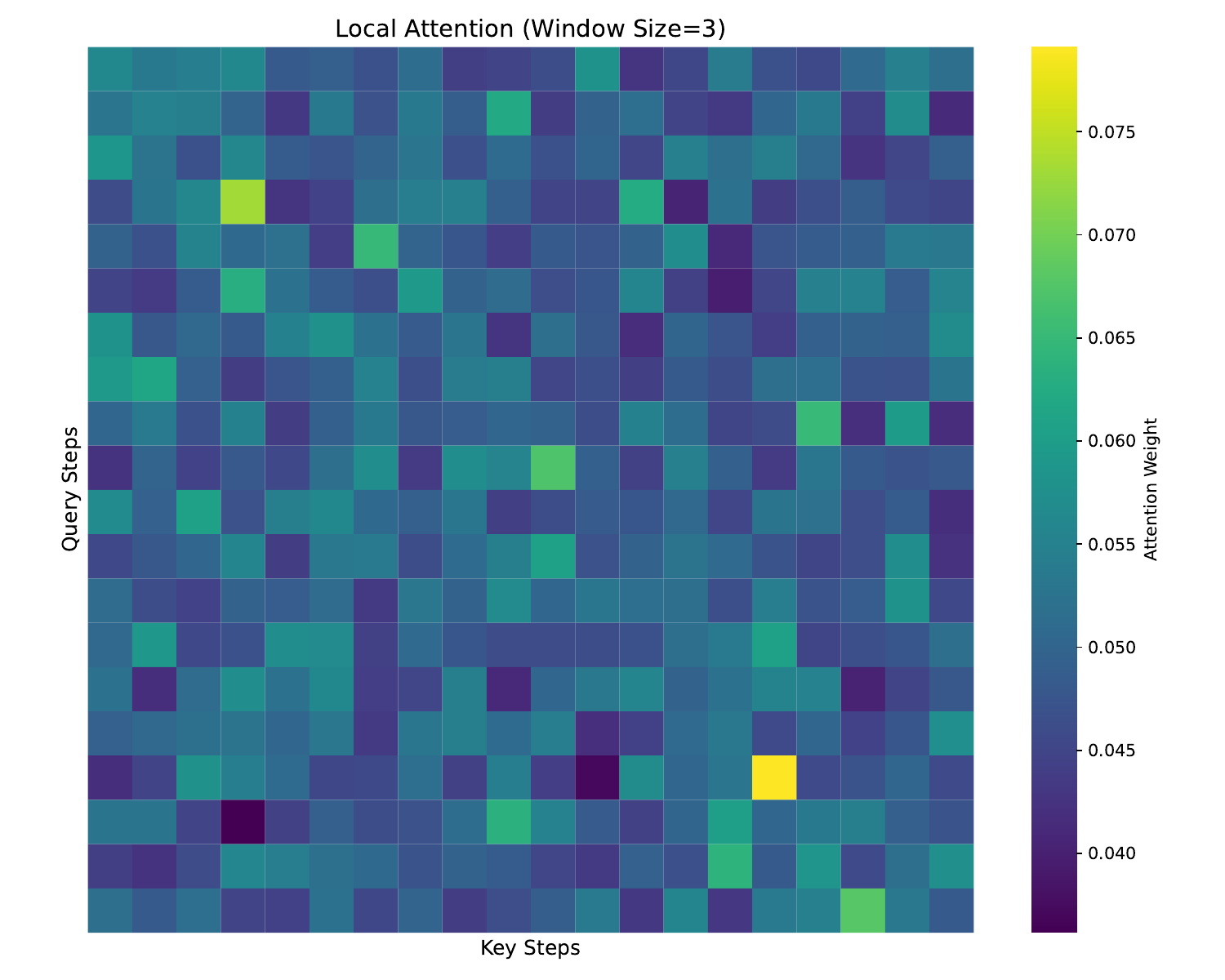}
  \subcaption{Cross Attention}
  \label{fig:ca}
\end{minipage}\hfill
\begin{minipage}[t]{0.32\textwidth}
  \centering
  \includegraphics[width=\linewidth,height=4cm]{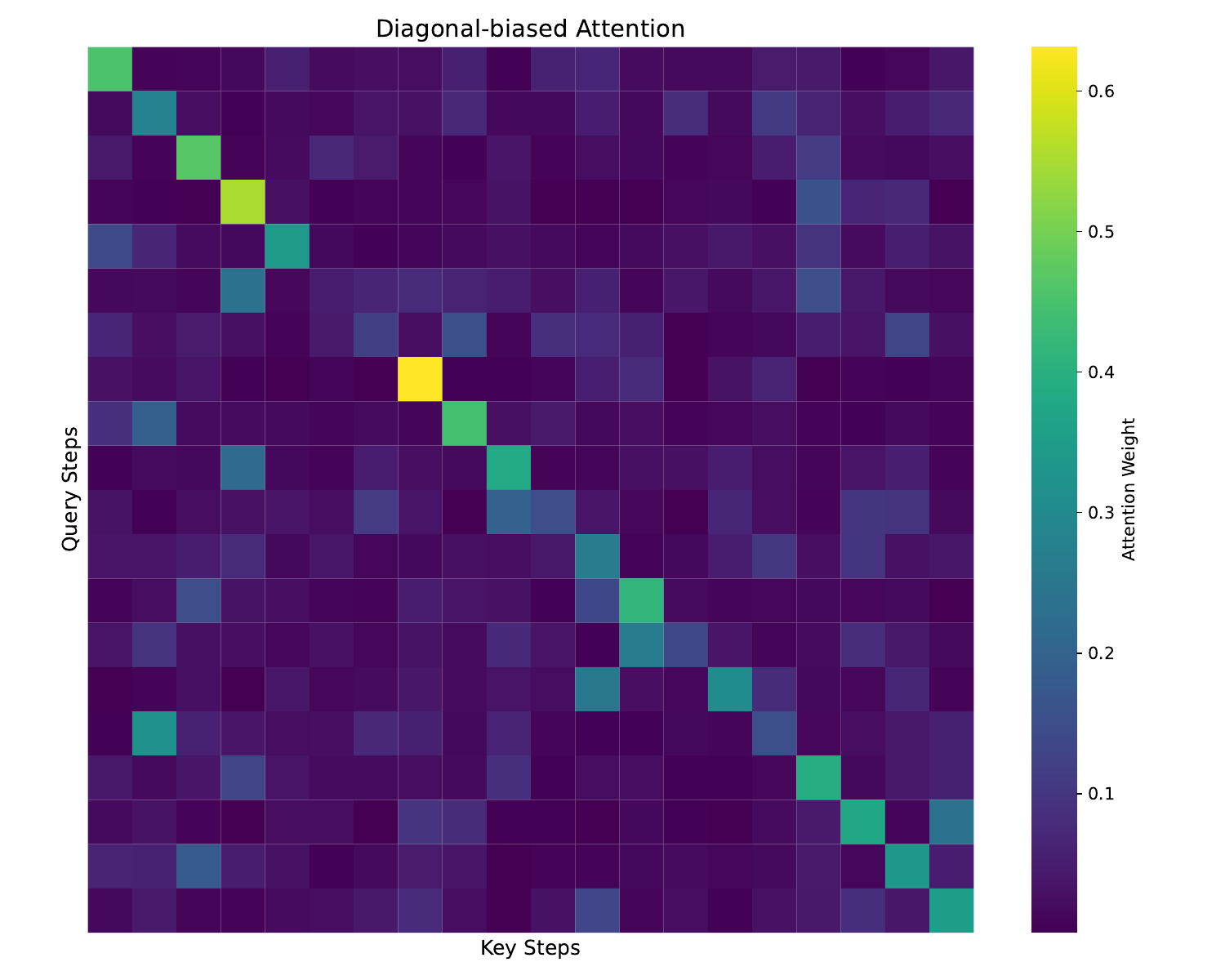}
  \subcaption{Hybrid Multi-Head Attention}
  \label{fig:hmha}
\end{minipage}

\vspace{0.5em}
\caption{Comparative Attention Heatmaps for Different Attention Mechanisms: An Analysis Based on the RR Interval Task in the MIMIC-IV Dataset.}
\label{fig:attention_heatmaps}
\end{figure}


Our ablation studies highlight key insights into ECG-MoE’s architectural design. 
As shown in left column of  Table~\ref{tab:unified_comparison_transposed} further shows that the rhythm-conditioned MoE achieves clear clinical advantages over conventional variants. 
By aligning expert routing with cardiac phase, ECG-MoE outperforms TimeMoE in RR-interval and potassium estimation while improving sex classification, confirming that phase-aware gating benefits morphology-sensitive tasks.

As shown in middle column of Table~\ref{tab:unified_comparison_transposed}, hybrid attention is vital for clinical precision. 
It consistently surpasses self- and cross-attention in RR-interval and age estimation and achieves the highest F1 across classification tasks. Figure~\ref{fig:attention_heatmaps} illustrates that hybrid attention effectively integrates local morphology and global rhythm context, producing interpretable, physiologically grounded predictions.

These findings show that ECG-MoE’s integration of multi-task periodicity learning, phase-aligned gating, and hybrid attention enables superior diagnostic accuracy and interpretability, confirming that ECG-specific inductive biases are essential for clinically reliable AI cardiology.

\section{Conclusion}

This study introduces ECG-MoE, a hybrid architecture that advances multi-task ECG analysis through specialized foundation integration. 
The model achieves SOTA performance across diverse tasks, reducing RR-interval MAE by 46.0\% compared to TEMPO and improving arrhythmia detection by 10.6\% over MOMENT, validating the effectiveness of its period-aware design and hybrid attention mechanisms. 
In addition, ECG-MoE establishes new efficiency benchmarks for clinical deployment, requiring only 8.2 GB of GPU memory and processing 14.7 samples per second, which enables real-time analysis on consumer-grade hardware and broadens accessibility in resource-limited settings. 
While the model demonstrates robust performance, challenges remain in potassium abnormality detection. 
Future efforts will focus on adaptive ensemble refinement and the incorporation of electrophysiological priors to enhance clinical decision support. 
The open-source release of ECG-MoE highlights the promise of domain-specific architectural innovations in advancing globally accessible, high-precision cardiac care.

\section{Acknowledgement}

This research was partially supported by the US National Science Foundation under Award Numbers 2319449, 2312502, and 2442172, as well as the US National Institute of Diabetes and Digestive and Kidney Diseases of the US National Institutes of Health under Award Number K25DK135913. We thank for the computing resources provided by the iTiger GPU cluster~\cite{sharif2025cultivating} supported by the NSF MRI program.


\bibliographystyle{unsrt}
\bibliography{ref}

\end{document}